\documentclass[sigconf]{acmart}

\usepackage{graphicx}
\usepackage{amsmath}
\usepackage{amssymb}
\usepackage{multirow}
\usepackage{amsmath}
\usepackage{amssymb}
\usepackage{subfigure}

\usepackage{balance}

\input{def.set}

\copyrightyear{2021}
\acmYear{2021}
\setcopyright{acmcopyright}
\acmConference[MM '21]{Proceedings of the 29th ACM International Conference on Multimedia}{October 20--24, 2021}{Virtual Event, China}
\acmBooktitle{Proceedings of the 29th ACM International Conference on Multimedia (MM '21), October 20--24, 2021, Virtual Event, China}
\acmPrice{15.00}
\acmDOI{10.1145/3474085.3475496}
\acmISBN{978-1-4503-8651-7/21/10}

\AtBeginDocument{%
  \providecommand\BibTeX{{%
    \normalfont B\kern-0.5em{\scshape i\kern-0.25em b}\kern-0.8em\TeX}}}

\begin{document}
\fancyhead{}

\title{Transferrable Contrastive Learning \\ for Visual Domain Adaptation}
\titlenote{This work was done when Yang Chen was a research intern at JD AI Research.}

\author{Yang Chen$^{1}$, Yingwei Pan$^{2}$, Yu Wang$^{2}$, Ting Yao$^{2}$, Xinmei Tian$^{1}$, Tao Mei$^{2}$}

\affiliation{%
  \institution{$^{1}$University of Science and Technology of China, Hefei, China}
        \country{$^{2}$JD AI Research, Beijing, China}
}
\email{cheny01@mail.ustc.edu.cn;{panyw.ustc, feather1014, tingyao.ustc}@gmail.com;xinmei@ustc.edu.cn;tmei@jd.com}

\renewcommand{\shortauthors}{Chen et al.}

\begin{abstract}
Self-supervised learning (SSL) has recently become the favorite among feature learning methodologies. It is therefore appealing for domain adaptation approaches to consider incorporating SSL. The intuition is to enforce instance-level feature consistency such that the predictor becomes somehow invariant across domains. However, most existing SSL methods in the regime of domain adaptation usually are treated as standalone auxiliary components, leaving the signatures of domain adaptation unattended. Actually, the optimal region where the domain gap vanishes and the instance level constraint that SSL peruses may not coincide at all. From this point, we present a particular paradigm of self-supervised learning tailored for domain adaptation, i.e., \textbf{T}ransferrable \textbf{C}ontrastive \textbf{L}earning (\textbf{TCL}), which links the SSL and the desired cross-domain transferability congruently. We find contrastive learning intrinsically a suitable candidate for domain adaptation, as its instance invariance assumption can be conveniently promoted to cross-domain class-level invariance favored by domain adaptation tasks. Based on particular memory bank constructions and pseudo label strategies, TCL then penalizes cross-domain intra-class domain discrepancy between source and target through a clean and novel contrastive loss. The free lunch is, thanks to the incorporation of contrastive learning, TCL relies on a moving-averaged key encoder that naturally achieves a temporally ensembled version of pseudo labels for target data, which avoids pseudo label error propagation at no extra cost. TCL therefore efficiently reduces cross-domain gaps. Through extensive experiments on benchmarks (Office-Home, VisDA-2017, Digits-five, PACS and DomainNet) for both single-source and multi-source domain adaptation tasks, TCL has demonstrated state-of-the-art performances.
\end{abstract}



\maketitle

\vspace{-0.25in}
\begin{figure}[h]
\includegraphics[width=1\linewidth]{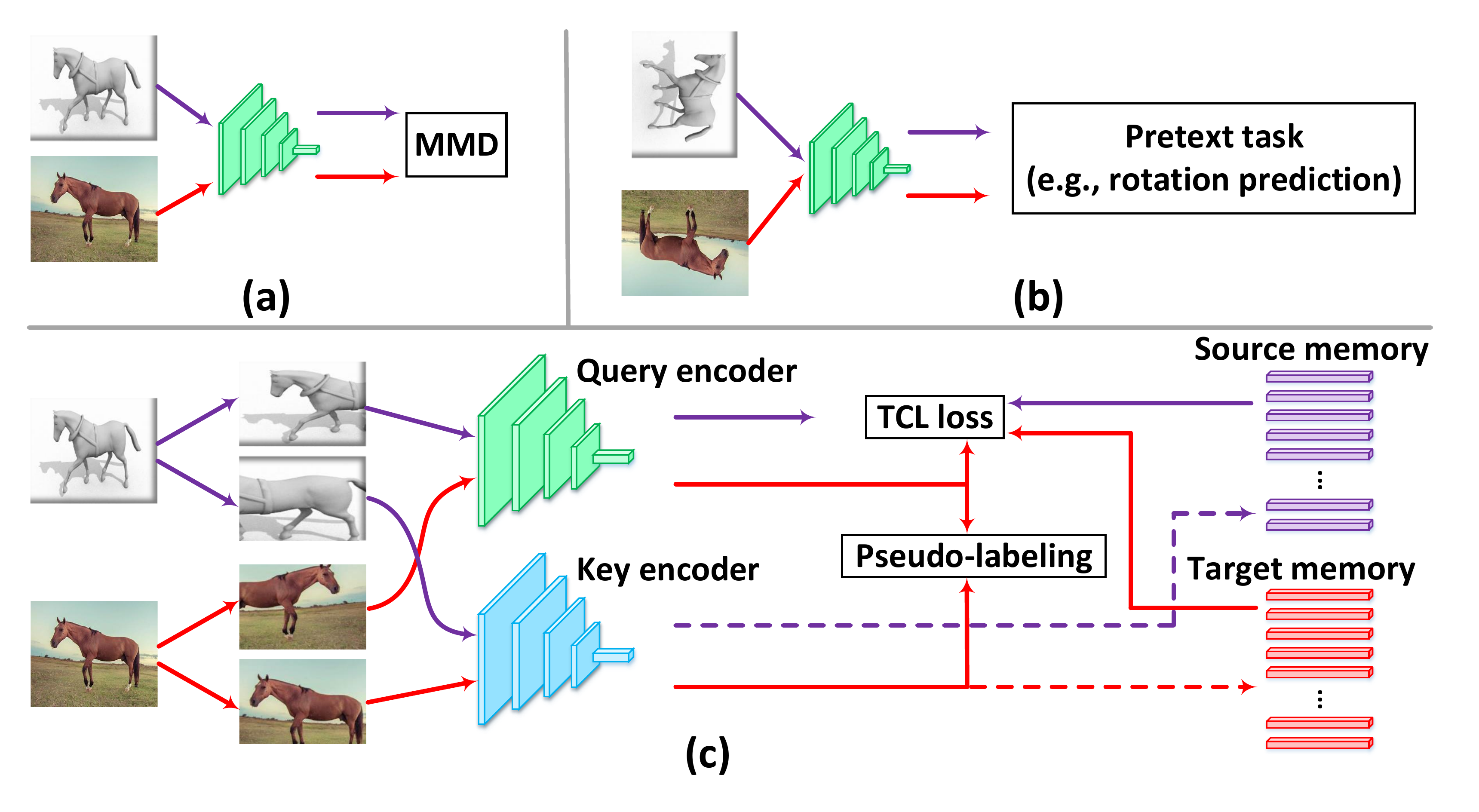}
\vspace{-0.3in}
	\caption{Comparison between (a) typical Unsupervised Domain Adaptation (UDA) that bridges domain gap via MMD, (b) Self-Supervised Learning (SSL) that strengthens feature discrimination through pretext tasks, e.g., rotation prediction, (c) Our Transferrable Contrastive Learning (TCL) which unifies SSL and UDA by integrating SSL pretext tasks and the crucial target of domain alignment in a contrastive manner.}
\vspace{-0.2in}
\label{fig:introduction}
\end{figure}

\section{Introduction}

Deep Neural Networks (DNN) \cite{he2016deep,li2021contextual} have shown powerful capability of feature learning when trained on large-scale datasets. Conventional supervised learning using DNN frameworks requires enormous annotated data via expensive and time-consuming manual labeling. This has constrained the scalability of DNN vision models, especially when the annotation is unavailable or only limited compared to a large number of parameters constructing the network. Unsupervised Domain Adaptation (UDA) aims to alleviate this problem by making the most out of the available labels and the learned knowledge obtained from a rich-resource domain (i.e., source domain with labeled instances), such that this information can be recycled and applied to understand the scarce-resource domain (i.e., target domain without label annotations). Due to the domain distributional shift between the source domain and target domain, much effort has been made to explicitly reduce domain shift by aligning source and target distributions. A typical example is \cite{long2015learning}, where the domain gap is reduced by minimizing the Maximum Mean Discrepancy (MMD) metric between representations of source and target instances (Figure \ref{fig:introduction}(a)).

Inspired by semi-supervised learning methods \cite{sajjadi2016regularization, tarvainen2017mean}, Self-Supervised Learning (SSL) has become an alternative that pioneers the way of UDA approaches \cite{sun2019unsupervised, carlucci2019domain, xiao2020self, french2017self}. The basic idea behind these SSL works is to reformulate the UDA task as a semi-supervised learning problem: to perform self-supervised pretext tasks (e.g., rotation prediction) using unlabeled target and labeled source data (Figure \ref{fig:introduction}(b)), and to perform supervised learning on labeled source data. Take for instance, \cite{sun2019unsupervised, xiao2020self} apply 2d rotation to training data, and the encoder has to learn useful semantic features to accomplish the rotation prediction task. Although these seminal SSL based UDA approaches have achieved state-of-the-art performances in comparison to conventional UDA frameworks in Figure \ref{fig:introduction}(a), existing SSL paradigms, unfortunately, tend to ignore the rich information and inherent distributional assumptions hidden behind the domain gaps. Since the domain gap assumption does not necessarily respect the various instance-level hypothesis that conventional SSL desires, there is still plenty of room to improve SSL frameworks when it is associated with UDA tasks.

The above considerations motivate us to tailor brand new SSL feature extraction recipes exclusively for UDA. Accordingly, we present Transferrable Contrastive Learning (TCL), a novel UDA algorithm that unifies and forties the strength of both SSL and UDA congruently. Our launching point is to refurbish conventional Contrastive Learning -- a recently prevailing SSL framework, such that our novel UDA distributional assumptions, useful prior knowledge in the UDA field, and long-existing old training tricks such as pseudo labeling can all be snugly plugged in with suitable modifications. To this end, we build domain-specific memories: source memory and target memory. The delicate design of these memories are of pivotal importance that validates TCL, and bridges SSL with domain adaptation tasks from there. We define source memory to track source domain instances' features along with their class information. We also have target memory that records target domain feature along with their ``pseudo-labels''. These particular slicing strategies of memories according to their domain and labels (or pseudo-labels) are essential for contrastive learning to rejuvenate in the context of UDA. The constructed domain-specific memories are aimed to offer a bed for later jointly modeling among heterogeneous pretext tasks (e.g., pseudo-labeling and class-level instance discrimination) while the domain gap is simultaneously reduced.

We summarize our contribution succinctly here: we propose a novel self-supervised learning paradigm called TCL. The TCL is a brand new UDA recipe to achieve both inter-class discriminativeness and cross-domain transferability of features through contrastive feature learning. TCL provides a tailored solution that focuses on the effective integration of contrastive learning and UDA via its associated pseudo labeling strategy and memory slicing policies. The TCL framework contrasts with any existing UDA algorithms that combine instance-level invariance rigidly as merely a standalone ``icing on the cake'' for UDA problems, and TCL discusses how the conventional contrastive learning could be recast exclusively to serve the UDA goals. We demonstrate that TCL achieves state-of-the-art empirical performances on five benchmarks, in regard to both single and multi-source domain adaptation tasks.

\section{Related Work}
\textbf{Unsupervised Domain Adaptation} methods can be generally categorized into four groups. The first group is the domain discrepancy based approach \cite{kang2019contrastive, long2015learning, long2016unsupervised, long2017deep, yan2017mind, yao2015semi}. Motivated by the theoretical bound proposed in \cite{ ben2010theory, ben2006analysis}, this direction aims at minimizing the discrepancy between the source and target domains. The adopted discrepancy loss varies across different statistic distances metrics (e.g., Maximum Mean Discrepancy (MMD) \cite{long2015learning}, Jensen-Shannon (JS) Divergence \cite{ganin2016domain} and Wasserstein distance \cite{shen2017wasserstein}). The second group usually resorts to adversarial learning, which aims to leverage adversarial learning to align two domains on either feature-level \cite{cui2020gradually, ganin2015unsupervised, long2018conditional, tzeng2017adversarial, li2019cycle, li2019joint} or pixel-level \cite{bousmalis2017unsupervised, ghifary2016deep, xu2020adversarial, chen2019mocycle}. The third group is the pseudo label based approach \cite{chen2019progressive, french2017self, gu2020spherical, pan2019transferrable, jing2020adaptively, cai2019exploring}. Take for example, \cite{chen2019progressive, pan2019transferrable} utilize pseudo labels to estimate class centers in the target domain, which are then used to match each class center across the source domain and target domain. The fourth direction is the recently emerging self-supervised learning based approaches \cite{sun2019unsupervised, carlucci2019domain, xiao2020self, french2017self}. In Jigsaw \cite{carlucci2019domain}, the permutation of images crops from each instance are predicted, so that the network improves its semantic learning capability for the UDA task.
In addition to single source domain UDA, works also consider the problem of multi-source domain adaptation task \cite{hoffman2018algorithms, peng2019moment, wang2020learning, xu2018deep, yang2020curriculum, zhao2018adversarial, pan2019multi}, which is a more generalized and practical scenario when annotated training data from multiple source domains are accessible.
For example, MDAN \cite{zhao2018adversarial} utilizes adversarial adaptation to induce invariant features for all pairs of source-target domains. M$^{3}$SDA \cite{peng2019moment} matches the statistical moments of all source domains and target domain to reduce the domain gap.

\begin{figure*}[th!]
\begin{center}
\vspace{-0.07in}
\includegraphics[width=1\linewidth]{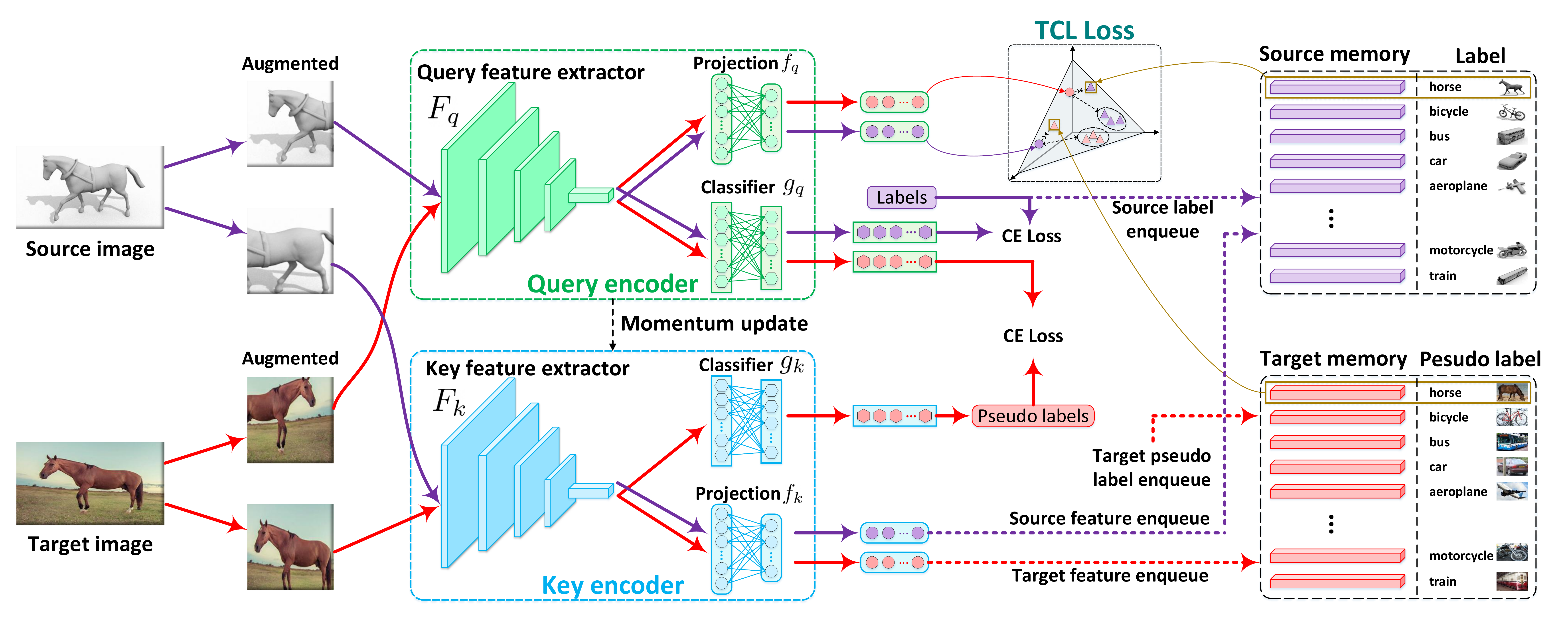}
\end{center}
\vspace{-0.2in}
	\caption{An overview of TCL. In addition to conventional supervised learning over labeled source query sample, pseudo-labeling is performed to supervise the prediction of target query sample via Cross-Entropy (CE) loss. TCL loss then minimizes the cross-domain discrepancy between query and positive keys belonging to the same class, while the assumed negative keys from different classes are stored in domain-specific memories. TCL loss is essentially a cross-domain class-level objective that naturally unifies both class-level instance discrimination and domain alignment objectives.}
\label{fig:framework}
\vspace{-0.16in}
\end{figure*}

\textbf{Contrastive Learning.}
Among the recent state-of-the-art self-supervised learning algorithms \cite{bachman2019learning, caron2020unsupervised, chen2020simple, chen2021empirical, he2020momentum, hjelm2018learning, oord2018representation, yao2021seco}, contrastive learning \cite{hadsell2006dimensionality, cai2020joint, lin2021core} has offered impressively superior performance, especially when used for pre-training tasks. The key idea of contrastive learning is to achieve \emph{instance-level discrimination} and \emph{invariance}, by pushing semantically distinct points away in the feature space, while pulling semantically nearby points closer. Inspired by NCE \cite{gutmann2010noise}, CPC \cite{oord2018representation} considers a softmax classifier to distinguish between individual instance classes.  As NCE theoretically justified, a large number of instances help to alleviate the learning problem. MoCo \cite{he2020momentum} thus further pushes the limit of contrastive training via the construction of a momentum updated memory bank, to store past old representations as more as possible. SimCLR \cite{chen2020simple}, SwAV \cite{caron2020unsupervised} and MoCo-v3 \cite{chen2021empirical} provide alternative training techniques that efficiently improve contrastive learning.

In this work, we seek to establish alternative novel symbiosis to combine contrastive learning with UDA task in a more congruent way. Our proposed TCL loss is inspired by contrastive learning but further goes beyond the \emph{instance level discrimination} in typical contrastive learning. In comparison to previous UDA methods that also utilize SSL, TCL aims to leverage and fortify the functional components of both class-level instance discrimination and domain alignment across different domains, and to achieve ``the whole is greater than the sum of its parts'' advantage out of the two.

\section{Approach}

In this work, we tailor self-supervised learning (e.g., contrastive learning) for UDA, by coupling SSL pretext tasks and the objective of domain alignment via a single framework in a contrastive fashion. An overview of our Transferrable Contrastive Learning (TCL) architecture is depicted in Figure \ref{fig:framework}.

\subsection{Preliminaries}
{\bf Contrastive Learning}. In the context of contrastive learning (e.g., MoCo \cite{he2020momentum}), each image $\bx_i$ is considered as an instance, specified by index $i$. A popular paradigm for contrastive learning is to first produce two randomly generated transformations of $\bx_i$ into query image $\bx^q_i$ and key image $\bx^k_i$, and then consider these two augmentations are coming from the same distribution, also specified by the instance $i$. The motivation of contrastive learning is to find some instance-invariant encoders so that $\bq_i=h_q(\bx^q_i)$ and $\bk^+_i=h_k(\bx^k_i)$ are as close as possible. In the meanwhile, the constructed $h_q$ and $h_k$ need to keep features $\bq_i$  and  $\bk^+_i$ discriminative against instances generated from other noise distribution. In this way, the contrastive loss aims to penalize dissimilarity within each positive pair ($\bq_i$, $\bk^+_i$). Given features $\bk^-_{i,j}, j \in [1,K]$ generated from other distributions, the encoders $h_q$ and $h_k$ are also expected to encourage discriminativeness within each negative pair ($\bq_i$ , $\bk^-_{i,j}$). A popular formulation to achieve this goal is to cast the contrastive learning into a classification problem, e.g., as in InfoNCE \cite{oord2018representation}, where the loss classifies positive pairs from $K$ negative pairs via a softmax~function:
\begin{equation}\scriptsize
\label{eq:cpc}
\mathcal{L}_{NCE}(\bq_i,\bk_i^{+})=-\log \frac{\exp(\left \langle \bq_i,\bk^{+}_i \right \rangle/\tau)} {\exp(\left \langle \bq_i,\bk_i^{+}\right \rangle/\tau)+\sum\limits_{j=1}^{K} \exp(\left \langle \bq_i,\bk_{i,j}^{-}\right \rangle/\tau)}.
\end{equation}
Here $\tau$ is the temperature parameter, and $\left \langle \bq_i,\bk^{+}_{i}\right \rangle$ denotes the inner product between $\bq_i$ and $\bk^+_i$. Different approaches also employ efficient policies for effectively generating negative pairs. For example, in MoCo, the key features are sequentially queued into a ``memory bank''. While $h_q$ is updated via backpropagation, the $h_k$ is only momentum updated according to the changes of $h_q$. In this way, the contrastive loss is able to read a large number of negative keys from the slowly evolving memory bank, so that the number of negative pairs are no longer constrained by the batch size.

{\bf Problem Formulation of UDA}. Consider a source domain having $N_s$ labeled samples $\mathcal{S}=\{(\bx^{s}_i, y^{s}_i)\}^{N_s}_{i=1}$ distributed across $C$ classes, where $y^{s}_i \in \{1,...,C\}$ is the label of the $i$th image $ \bx^{s}_i$. In the meanwhile, assume there is also a target domain with $N_t$ number of unlabeled samples $\mathcal{T}=\{\bx^{t}_i\}^{N_t}_{i=1}$ ranging over the same $C$ as in the source domain, without any annotations though. The goal of UDA is to train a network by exploiting labeled data from source domain $\mathcal{S}$ and unlabeled data from $\mathcal{T}$, so that the obtained model may also be well adapted to the target domain on prediction tasks, even if there is evident and non-negligible domain gap present between the distribution of $\mathcal{S}$ and $\mathcal{T}$.

{\bf Notations.} We firstly generate two augmentations of each source input $\bx^s_i$, i.e., respectively into $\bx^{s,q}_i$ and $\bx^{s,k}_i$. Analogically, we also generate two augmentations of each target data $\bx^t_i$, denoted as $\bx^{t,q}_i$ and $\bx^{t,k}_i$. According to the standard paradigms in contrastive learning, we implement a query feature extractor  $\bz^{s,q}_i=F_q(\bx^{s,q}_i, \btheta_q)$, followed by an additional query projection $\bq^s_i=f_q(\bz^{s,q}_i,\bbeta_q)\in \mathbb{R}^d$, where $\bq^s_i$ is the source query vector of the $i$ instance. Similarly, we deploy another key feature extractor $\bz^{s,k}_i=F_k(\bx^{s,k}_i, \btheta_k)$, followed by a key projection $\bk^s_i=f_k(\bz^{s,k}_i,\bbeta_k)\in \mathbb{R}^d$, where $\bk^s_i$ is the source key vector of the $i$ instance. In parallel, we also require target query and target key features are generated via exactly the same encoders: i.e., $\bq^t_i=f_q(\bz^{t,q}_i, \bbeta_q)\in \mathbb{R}^d$, where $\bz^{t,q}_i=F_q(\bx^{t,q}_i, \btheta_q)$;  and $\bk^{t}_i=f_k(\bz^{t,k}_i,\bbeta_k)\in \mathbb{R}^d$, where $\bz^{t,k}_i=F_k(\bx^{t,k}_i, \btheta_k)$. We also respectively define query classifier $g_q(\cdot, \bpsi_q, c )$ and key classifier $g_k(\cdot, \bpsi_k, c )$, where the symbol ``$\cdot$'' corresponds to the relevant input depending on the context. Accordingly, the group of query/key feature extractor, projection and classifier can be naturally treated as the query/key encoder. During training, the parameters of query encoder ($\bTheta_q=\{ \btheta_q, \bbeta_q, \bpsi_q\}$) are updated via backpropagation according to the loss, whereas the parameters of key encoder ($\bTheta_k=\{ \btheta_k, \bbeta_k, \bpsi_k\}$) are only momentum updated. We defer the parameter update details to the later Section \ref{sec.optimization}.

\subsection{TCL framework}
In short, TCL capitalizes on particular class-level invariance assumption and memory bank slicing policy to tackle the UDA problems. These hypotheses are expected to leverage the benefits from both domain adaptation techniques and self-supervised feature learning in a coupled manner. The intuition here is that, we still are able to construct positive and negative pairs across domains based on true/tentative label predictions over source/target data, so that intra-class similarity and domain alignment are learned in a reliable {\em contrastive} way. However, since the conventional contrastive learning assumptions purely aim at instance-level invariance, a rigid combination of the two (i.e., UDA and contrastive learning) does not necessarily reduce the degree of freedom of the problem when searching the parameter regions that minimizes the domain gaps. We fix this issue by introducing TCL.

{\bf Pseudo-labeling in Target}. To start with, we define our specific pseudo-labeling method for the unlabeled target domain. The motivation behind this is to promote instance-level invariance intrinsically favored by contrastive learning further to the cross-domain class-level invariance desired by UDA.
Let the probability $p(c |\bz^{s,q}_i)$ denote the predicted classification distribution across $c \in [1,...C]$ classes, given the source query feature $\bz^{s,q}_i$. We minimize the standard cross-entropy loss between the prediction and the ground truth label distribution for each source sample:
\begin{equation}\small
\label{eq:sourceentropy}
\mathcal{L}_{src} =-\sum \limits_{(x^s_i, y^s_i)}{y^s_i}\log{p(y^s_i=c |\bz^{s,q}_i)}, 
\end{equation}
where the class prediction is obtained via classifier $p(y^s_i=c |\bz^{s,q}_i) = g_q(\bz^{s,q}_i, \bpsi_q, c )$, and $\bpsi_q$ is the classifier parameter. Conventionally, the obtained classifier $g_q(\cdot, \bpsi_q, c )$ should be able to classify unseen test data from the same distribution as $\mathcal{S}$. In the context of UDA though, it is necessary that $g_q(\cdot, \bpsi_q, c )$ is also exposed to target data $\mathcal{T}$, so that the classifier becomes invariant and robust against domain distributional discrepancy.

According to the discussion in the seminal work \cite{ben2010theory}, the domain discrepancy between $\mathcal{S}$ and $\mathcal{T}$ must satisfy the ideal joint hypothesis, so that the domain adaptation problem itself is applicable. If this hypothesis performs poorly, we cannot expect to learn any good target classifier by minimizing source error. Based on this principle, the classification error from directly implementation the classifier on target data is rigorously upper bounded.

The ideal joint hypothesis therefore reassures us that provided $g_q(\cdot, \bpsi_q, c )$ is accurate on source data $\mathcal{S}$, the error from implementing $g_q(\cdot, \bpsi_q, c )$ directly on target data likely remains acceptable to some extent. Otherwise, the target domain probably is too distinct to adapt for, and any adaptation approaches will be irrelevant. Based on this hypothesis, we label each target data tentatively with the predication $\hat{y_i^t}$ through the key classifier $g_k(\cdot, \bpsi_k, c )$:
\begin{align}
&\hat{y_i^t}=\argmax_c g_k(\bz^{t,k}_i, \bpsi_k, c ),
\label{eq:psedolabel}
\end{align}
where $\bz^{t,k}_i=F_k(\bx^{t,k}_i, \btheta_k)$. During each iteration of training, the classifier $g_q(\cdot, \bpsi_q, c )$ is updated via backpropogation, while  $g_k(\cdot, \bpsi_k, c )$ is only momentum updated.

Note the unique training mechanism of classifier $g_k(\cdot, \bpsi_k, c )$ is critical for our TCL approach to succeed. Mathematically speaking, the momentum updated $g_k(\cdot, \bpsi_k, c )$ is equivalently a temporally ensembled version of $g_q(\cdot, \bpsi_q, c )$. This implicit averaging strategy effectively reduces the pseudo label error propagation owing to the presence of the domain gap, and efficiently stabilizes the classification on target data. Accordingly, we assign the predicted $\hat{y_i^t}$ value to be the class ``pseudo label'' of sample $\bx^{t,q}_i$. The standard cross entropy loss $\mathcal{L}_{tar}$ on target data is then computed as the target classification error, iff. the prediction $p( \hat{y_i^t}=c |\bz^{t,k}_i) = g_k(\bz^{t,k}_i, \bpsi_k, c )$ exceeds a threshold $\rho$ ($\rho=0.95$):
\begin{equation}
\label{eq:losstarget}
\small
\mathcal{L}_{tar} =-\sum \limits_{x^t_i}\P({ p(\hat y_i^t = c |\bz^{t,k}_i)}>\rho)\log{p(\hat y_i^t=c |\bz^{t,q}_i)}.
\end{equation}
Here, the indicator function $\P(x)=1$ if condition $x$ holds, otherwise $\P(x)=0$. Please pay attention to Eq. (\ref{eq:losstarget}), where the pseudo label prediction $p(\hat y_i^t=c |\bz^{t,k}_i)$ is computed given the target {\em key features} $\bz^{t,k}_i$, but used to supervise the target {\em query features} update via prediction $p(\hat y_i^t=c |\bz^{t,q}_i)$. This echoes our motivation that~classifier $g_k(\cdot, \bpsi_k, c )$ is a temporally ensembled version of~$g_q(\cdot, \bpsi_q, c )$, whose parameter evolution is more reliable and stable than conventional pseudo labeling strategy.

{\bf Key Encoder and Domain-specific Memories}. Our memory bank definition is intentionally tailored to serve UDA task. Recall that during each forward pass, we have obtained key features $\bk^s_i=f_k(\bz^{s,k}_i, \bbeta_k)$. We enqueue each $(\bk^s_i, y^s_i)$ pair sequentially into each of the domain specific {\em source memory bank} ($D$ number of banks if multiple sources). Following \cite{kang2019contrastive}, the spherical K-means is also adopted here to perform the clustering on target samples, which helps further refine the pseudo label prediction $\hat y^t_i$. We then sequentially enqueue each batch of target key feature and the corresponding pseudo label $(\bk^t_i, \hat y^t_i)$ into the {\em target memory bank}. In regard of memory bank updates, TCL shares similar spirit of \cite{he2020momentum}: As the training proceeds, the oldest batch of keys in each target/source memory bank are removed, and the current batch of keys are enqueued. But in comparison to conventional contrastive learning, the main goal of TCL is rather to efficiently log both the feature simultaneously with their label/psudo label information in a pairwise manner, which eases our implementation of TCL loss as follows.

{\bf TCL Loss: from Instance-level to Cross-domain Class-level Invariance}. Having all of the notations clarified in the previous sections, we eventually arrive at the transferable contrastive learning loss: TCL loss, our essential adaptation mechanism. The introduction of TCL loss is expected to naturally leverage the heterogeneous self-supervised learning tasks, i.e., instance-level invariance into cross-domain class-level invariance that favored by both contrastive learning \cite{he2020momentum} and UDA problem itself.

Formally, in a training batch, we define all source query $\bq^s_i$ that belong to some specific class $c$ as positive queries to class $c$. We also consider all the target samples $\bk^t_i$ currently enqueued in the target memory bank with pseudo label $\hat y_i^t=c$ as positive keys of class $c$. All of the remaining keys in the target memory bank are treated as negative keys to this specific class $c$. Mathematically, we yield:
\begin{equation}
\label{eq:lossst}
\scriptsize
\mathcal{L}({s,t})= - \sum_{c} \sum_{i,j} \P( y^s_i=c, \hat y^t_j=c)\log \frac{ \exp(\left \langle \bq^s_i,\bk_j^{t}\right \rangle/\tau)  }     { \exp(\left \langle \bq^s_i,\bk_j^{t}\right \rangle/\tau)  +\sum_{\ell \neq j, \hat y^t_\ell\neq c} \exp(\left \langle \bq^s_i,\bk_\ell^t\right \rangle/\tau)  },
\end{equation}
where class category $c \in [1, C]$ ranges over all existing classes in the current batch. Analogically, we also construct a $\mathcal{L}({t,s})$ that is symmetric to the definition of $ \mathcal{L}({s,t})$:
\begin{equation}
\label{eq:lossts}
\scriptsize
\mathcal{L}({t,s})=-\sum_{c} \sum_{i,j}\P( \hat y^t_i=c, y^s_j=c)\log \frac{ \exp(\left \langle \bq^t_i,\bk_j^{s}\right \rangle/\tau)  }{ \exp(\left \langle \bq^t_i,\bk_j^{s}\right \rangle/\tau)  +\sum_{\ell \neq j,  y^s_\ell\neq c} \exp(\left \langle \bq^t_i,\bk_\ell^s\right \rangle/\tau)}.
\end{equation}
The TCL loss boils down to:
\begin{equation}
\label{eq:tcl}
\small
\mathcal{L}_{tcl}=\mathcal{L}({t,s})+\mathcal{L}({s,t}).
\end{equation}

It is transparent now that TCL is a cross-domain class-level contrastive loss, an effective extension of the instance-invariance assumption that snugly bridges contrastive learning with UDA. In comparison to conventional {\em{instance level}} contrastive learning as in Eq, (\ref{eq:cpc}), the motivation behind Eq. (\ref{eq:tcl}) is quite straightforward: contrastive loss penalizes incompatibility of each {\em cross-domain} positive pairs $(\bq^t_i,\bk_j^{s} )$ and $(\bq^s_i,\bk_j^{t})$ that most likely fall into the same category, given the predicted pseudo label on target samples. Eq. (\ref{eq:tcl}) also effectively pushes away {\em cross-domain} negative pairs if these samples are believed to be coming from distinct class categories. Accordingly, such class-level contrastive learning naturally erases the intra-class feature variance between source and target, while inter-class feature discrimination is further improved.

{\bf Extension to Multi-source Domain Adaptation}. An immediate extension of our proposal is to jointly apply Eq. (\ref{eq:tcl}) across multiple domains. Consider the scenario where we have a dataset composed of $M$ distinct labeled source domains: $\mathcal{S}_1$,$\mathcal{S}_2$,...,$\mathcal{S}_m$,...,$\mathcal{S}_M$ distributed across classes $c\in [1,C]$. Our goal becomes to best exploit all of these diverse sources and the associated annotations simultaneously so that we can improve the prediction on unlabeled target data $\mathcal{T}$.  One extra bonus out of the proposed TCL loss is that, Eq. (\ref{eq:tcl}) can be conveniently plugged into this multi-source scenario with least modifications:
\begin{equation}
\small
\label{eq:tclM}
\mathcal{L}_{tcl-M}=\sum_{m,n, m\neq n}^M \mathcal{L}({s_m,s_n})+  \sum_{m}^M \mathcal{L}({s_m,t}) + \sum_{m}^M \mathcal{L}({t,s_m}).
\end{equation}
Here, we introduce an extra term $\mathcal{L}({s_m,s_n})$ that takes into account the {\em cross source-source domain} feature correlations. Loss $\mathcal{L}({s_m,s_n})$ basically retains the same form of $\mathcal{L}(s,t)$ as in Eq. (\ref{eq:lossst}), and only differs in that the positive/negative pair construction for $\mathcal{L}({s_m,s_n})$ is completely based on ground truth annotation of {\em cross source-source domain} data, instead of the utility of pseudo labels in $\mathcal{L}(s,t)$. Correspondingly, we leverage the cross-domain class-level correlation for each specific class $c$ by capitalizing on all of the annotations available from multiple sources. This potentially imposes stronger cross-domain feature invariance for each class, so that the pseudo label predication on $\mathcal{T}$ becomes more reliable.

{\bf Overall Objective.} \label{sec.optimization}
To summarize, for multi-source adaption, we minimize the overall loss $\mathcal{L}_{total-M}$ for each training batch:
\begin{equation}
\label{eq:finallossmulti}
\small
\mathcal{L}_{total-M}=\mathcal{L}_{src} +\mathcal{L}_{tar} + \lambda \mathcal{L}_{tcl-M}.
\end{equation}
The hyperparameter $\lambda$ trades-off the impact of ${L}_{tcl-M}$ against the classification loss $\mathcal{L}_{src}$ and $\mathcal{L}_{tar}$. Note the $\mathcal{L}_{tcl-M}$ loss reduces to the plain form of $\mathcal{L}_{tcl}$ if and only if there is only a single labeled source domain of data considered., i.e., when $M=1$. This leads to our overall loss $\mathcal{L}_{total-S}$ for single-source adaptation problem:
\begin{equation}
\label{eq:finallosssingle}
\small
\mathcal{L}_{total-S}=\mathcal{L}_{src} +\mathcal{L}_{tar} + \lambda \mathcal{L}_{tcl}.
\end{equation}
During the training, the parameters in query and key encoders are updated in order to optimize Eq. (\ref{eq:finallossmulti}) or Eq. (\ref{eq:finallosssingle}) depending on the actual task. Specifically, the parameters of query encoder $\bTheta_q=\{ \btheta_q, \bbeta_q, \bpsi_q\}$ are updated via traditional backpropagation, whereas the parameters of key encoder $\bTheta_k=\{ \btheta_k, \bbeta_k, \bpsi_k\}$ is momentum updated as:
\begin{equation}
\label{eq:momentumupdate}
\small
\bTheta_k(it) = \alpha \cdot \bTheta_k(it-1)+(1-\alpha) \cdot \bTheta_q(it-1),
\end{equation}
where $it$ indicates the $it^{th}$ training iteration and $\alpha \in \left[0, 1 \right)$ is a momentum cofficient.

\begin{table*}[!ht]
\caption{Performance comparison with the state of arts on \emph{Office-Home} dataset.}

\vspace{-0.1in}
	\begin{center}
		\resizebox{0.96\textwidth}{!}{
			\begin{tabular}{l c c c c c c c c c c c c c}
				\toprule
				Method & Ar$\rightarrow$Cl & Ar$\rightarrow$Pr & Ar$\rightarrow$Rw & Cl$\rightarrow$Ar &Cl$\rightarrow$Pr & Cl$\rightarrow$Rw & Pr$\rightarrow$Ar & Pr$\rightarrow$Cl & Pr$\rightarrow$Rw & Rw$\rightarrow$Ar & Rw$\rightarrow$Cl & Rw$\rightarrow$Pr    & Avg  \\
				
				\midrule
				Source Only     & 34.9      & 50.0     & 58.0      & 37.4      & 41.9      & 46.2     & 38.5     & 31.2     & 60.4     & 53.9     & 41.2     & 59.9 & 46.1 \\
				
				DAN \cite{long2015learning}                 & 43.6     & 57.0     & 67.9      & 45.8      & 56.5      & 60.4     & 44.0     & 43.6     & 67.7     & 63.1     & 51.5     & 74.3  & 56.3 \\
				
				DANN \cite{ganin2016domain}     & 45.6     & 59.3     & 70.1      & 47.0      & 58.5      & 60.9     & 46.1     & 43.7     & 68.5     & 63.2     & 51.8      & 76.8 & 57.6 \\
				
				JAN \cite{long2017deep}                 & 45.9     & 61.2     & 68.9      & 50.4      & 59.7      & 61.0     & 45.8     & 43.4     & 70.3     & 63.9     & 52.4      & 76.8 & 58.3 \\
				
				SE \cite{french2017self}  & 48.8 & 61.8 & 72.8 & 54.1 & 63.2 & 65.1 & 50.6 & 49.2 & 72.3 & 66.1 & 55.9 & 78.7 & 61.5 \\
				
				DWT-MEC \cite{roy2019unsupervised} & 50.3 & 72.1 & 77.0 & 59.6 & 69.3 & 70.2 & 58.3 & 48.1 & 77.3 & 69.3 & 53.6 & 82.0 & 65.6 \\
				
				TAT \cite{liu2019transferable} & 51.6 & 69.5 & 75.4 & 59.4 & 69.5 & 68.6 & 59.5 & 50.5 & 76.8 & 70.9 & 56.6 & 81.6 & 65.8 \\
				
				SAFN \cite{xu2019larger} & 52.0 & 71.7 & 76.3 & 64.2 & 69.9 & 71.9 & 63.7 & 51.4 & 77.1 & 70.9 & 57.1 & 81.5 & 67.3 \\
				
				TADA \cite{wang2019transferable}       & 53.1 & 72.3 & 77.2 & 59.1 & 71.2 & 72.1 & 59.7 & 53.1 & 78.4 & 72.4 & 60.0 & 82.9 & 67.6 \\
				
				SymNets \cite{zhang2019domain} & 47.7 & 72.9 & 78.5 & 64.2 & 71.3 & 74.2 & 64.2 & 48.8 & 79.5 & 74.5 & 52.6 & 82.7 & 67.6 \\
				
				MDD \cite{zhang2019bridging}         & 54.9 & 73.7 & 77.8 & 60.0 & 71.4 & 71.8 & 61.2 & 53.6 & 78.1 & 72.5 & 60.2 & 82.3 & 68.1 \\
				
                SSDA \cite{xiao2020self} & 51.7 & 69.0 & 75.4 & 60.4 & 70.3 & 70.7 & 57.7 & 53.3 & 78.6 & 72.2 & 59.9 & 81.7 & 66.7 \\

				SRDC \cite{tang2020unsupervised} & 52.3 & 76.3 & 81.0 & 69.5 & 76.2 & 78.0 & 68.7 & 53.8 & 81.7 & 76.3 & 57.1 & 85.0 & 71.3 \\
                GVB-GD \cite{cui2020gradually} &57.0 &74.7 &79.8 &64.6 &74.1 &74.6 &65.2 &55.1 &81.0 &74.6 &59.7 &84.3 &70.4 \\
                RSDA \cite{gu2020spherical} &53.2 &77.7 &81.3 &66.4 &74.0 &76.5 &67.9 &53.0 &82.0 &75.8 &57.8 &85.4  & 70.9\\
				\midrule
				TCL (ResNet-50) & \textbf{59.4} & \textbf{78.8} & \textbf{81.6} & \textbf{69.9} & \textbf{76.9} & \textbf{78.9} & \textbf{69.2} & \textbf{58.7} & \textbf{82.4} & \textbf{76.9} & \textbf{62.7} & \textbf{85.6} & \textbf{73.4} \\
				\bottomrule
			\end{tabular}
		}
	\end{center}
\vspace{-0.1in}
	\label{table:results_officehome}
\end{table*}

\begin{table*}[ht]
\vspace{-0.0in}
\setlength\tabcolsep{10.2pt}
\caption{Performance comparison with the state of arts on \emph{VisDA-2017} dataset.}
\vspace{-0.1in}
\begin{center}
\resizebox{0.96\textwidth}{!}{
\begin{tabular}{ l  c  c   c   c   c   c  c c c c c c c}
\toprule
Method & plane & bicycle & bus & car & horse & knife & mcycl & person & plant & sktbrd & train & truck & Avg \\
\midrule
Source Only   &  72.3 & 6.1 & 63.4 & \textbf{91.7} & 52.7 & 7.9 & 80.1 & 5.6  & 90.1 & 18.5 & 78.1 & 25.9 & 49.4 \\
RevGrad \cite{ganin2015unsupervised} & 81.9 & 77.7 & 82.8 & 44.3 & 81.2 & 29.5 & 65.1 & 28.6 & 51.9 & 54.6 & 82.8 & 7.8 & 57.4 \\
DAN \cite{long2015learning}   & 68.1 & 15.4 & 76.5 & 87.0 & 71.1 & 48.9 & 82.3 & 51.5 & 88.7 & 33.2 & 88.9 & 42.2 & 62.8 \\
JAN \cite{long2017deep}       & 75.7 & 18.7 & 82.3 & 86.3 & 70.2 & 56.9 & 80.5 & 53.8 & 92.5 & 32.2 & 84.5 & 54.5 & 65.7 \\
MCD \cite{saito2018maximum}   & 87.0 & 60.9 & 83.7 & 64.0 & 88.9 & 79.6 & 84.7 & 76.9 & 88.6 & 40.3 & 83.0 & 25.8 & 71.9 \\
ADR \cite{saito2017adversarial}   & 87.8 & 79.5 & 83.7 & 65.3 & 92.3 & 61.8 & 88.9 & 73.2 & 87.8 & 60.0 & 85.5 & 32.3 & 74.8 \\

TPN \cite{pan2019transferrable}  &93.7 &85.1 &69.2 &81.6 &93.5 &61.9 &89.3 &81.4 &93.5 &81.6 &84.5 &49.9 &80.4\\

SE \cite{french2017self}  &  95.9 & 87.4   & 85.2 &  58.6  & 96.2  &  95.7  & 90.6 & 80.0 & 94.8 & 90.8 & 88.4 & 47.9 & 84.3 \\

SE-CC \cite{pan2020exploring}  &96.3 &86.5 &82.4 &81.3 &96.1 &97.2 &91.2 &84.7 &94.4 &94.1 &88.3 &53.4 &87.2\\

CAN \cite{kang2019contrastive}     & 97.0 & 87.2 & 82.5 & 74.3 & 97.8 & 96.2 & 90.8 & 80.7 & 96.6 & 96.3 & 87.5 & 59.9 & 87.2 \\

CAN\cite{kang2019contrastive}+ A$^{2}$LP \cite{zhang2020label}  &\textbf{97.5} &86.9 &83.1 &74.2 &\textbf{98.0} &\textbf{97.4} &90.5 &80.9 &96.9 &\textbf{96.5} &89.0 &60.1 &87.6 \\

\midrule

TCL (ResNet-101)    & 97.3 &\textbf{91.5}  &\textbf{85.9}& 73.9 & 96.6 & 97.1 & \textbf{93.6} & \textbf{85.1} & \textbf{97.0} & 96.1 & \textbf{89.9} & \textbf{70.9} & \textbf{89.6} \\
\bottomrule
\end{tabular}}
\end{center}
\vspace{-0.1in}
\label{table:cls-visda}
\end{table*}

\section{Experiments}\label{sec:exp}

\textbf{Datasets.} We evaluate the domain adaptation approaches on the following 5 datasets:
\textbf{\em A) Office-Home} \cite{venkateswara2017deep} is often considered as a challenging dataset for visual domain adaptation tasks, which consists of $15,500$ images distributed across $C=65$ classes. The Office-Home dataset is a composition dataset including four distinct domains: Artistic (\textbf{Ar}), Clip Art (\textbf{Cl}), Product  (\textbf{Pr}) and Real-World (\textbf{Rw}). Following \cite{french2017self,ganin2016domain,long2017deep}, we consider single source adaptation on this data. The adaptation directions range over all possible permutations of 2 out of 4 domains. For example, direction Ar$\rightarrow$Cl means domain Ar is labeled source domain and  Cl is the unlabeled target domain.
\textbf{\em B) VisDA-2017} \cite{peng2017visda} is a large scale dataset containing $152,409$ synthetic images (Syn) of $12$ classes in the source training set and $55,400$ real images (Real) from MS-COCO as the target domain validation set. We only consider the single adaption direction Syn$\rightarrow$Real, following the state-of-the-art approaches. \textbf{\em C) Digits-five} \cite{xu2018deep} is a popular benchmark for multi-source domain adaptation (MSDA), which consists of $10$ classes of digit images respectively sampled from five different datasets, including \emph{MNIST} (\textbf{mt}) \cite{lecun1998gradient}, \emph{MNIST-M} (\textbf{mm}) \cite{ganin2015unsupervised}, \emph{SVHN} (\textbf{sv}) \cite{netzer2011reading}, \emph{USPS} (\textbf{up}) \cite{hull1994database} and \emph{Synthetic Digits} (\textbf{sy}) \cite{ganin2015unsupervised}. These five datasets essentially represent five distinct domains. Following \cite{peng2019moment,xu2018deep}, we construct training/test dataset for \emph{MNIST}, \emph{MNIST-M}, \emph{SVHN} and \emph{Synthetic Digits} by randomly sampling $25,000$ training images/$9,000$ test images from each of the corresponding domain. For \emph{USPS}, we used their complete training and test datasets. The adaptation direction on Digits-five is defined as follows. Each individual domain rotates its role as the unlabeled target domain, while the remaining 4 domains are regarded as multiple labeled source domains. Take for instance $\rightarrow$ \textbf{sv} means dataset {\em sv} is the target dataset, while all of the remaining domains are considered as the labeled source domains. Therefore, there are in total 5 adaptation directions under this {\em Digits-five} dataset.
\textbf{\em D) PACS} \cite{li2017deeper} is another popular benchmark for MSDA, which is composed of four domains (Art, Cartoon, Photo and Sketch). Each domain includes samples from $7$ different categories, including a total of $9,991$ samples.
\textbf{\em E) DomainNet} \cite{peng2019moment} is a more challenging benchmark frequently used for evaluations on MSDA. The \emph{DomainNet} dataset contains samples from $6$ domains: Clipart (\textbf{clp}), Infograph (\textbf{inf}), Painting (\textbf{pnt}), Quickdraw (\textbf{qdr}), Sketch (\textbf{skt}), and Real (\textbf{rel}). Each domain has $345$ classes, and the dataset has $596,010$ images in total. We use one domain as target and the remaining as sources, as the same setting of \cite{peng2019moment, yang2020curriculum}.

\textbf{Implementation Details.}
For fair comparison with SOTA approaches across different datasets, we utilize the commonly adopted ImageNet pre-trained ResNet-50/101/18/101 as feature extractor (\emph{i.e.,} $F_q/F_k$) for \emph{Office-Home}, \emph{VisDA-2017}, \emph{PACS} and \emph{DomainNet}, respectively. For \emph{Digits-five}, the feature extractor is composed of three \emph{conv} layers and two \emph{fc} layers, which is also exploited in \cite{peng2019moment}. We adopt task-specific \emph{fc} layer to parameterize the classifier $g_q/g_k$. The projection $f_q/f_k$ is implemented by a \emph{fc} layer with an output dimension of $256$. We finetune from the pre-trained layers and train the newly added layer, where the learning rate of the latter is $10$ times that of the former. We use mini-batch SGD with momentum of $0.9$ to train the network for all experiments. The initial learning rate is $0.001$ for the convolutional layers and $0.01$ for the newly added layers. We follow the same learning rate schedule as in \cite{ganin2015unsupervised, long2015learning, long2017deep} and \cite{peng2019moment} for single-source and multi-source UDA, respectively. Inspired by \cite{sohn2020fixmatch}, we use a standard flip-and-shift augmentation strategy for $\mathcal{L}_{src}$ and adopt RandAugment \cite{cubuk2020randaugment} for $\mathcal{L}_{tar}$ and $\mathcal{L}_{tcl}$($\mathcal{L}_{tcl-M}$).
The temperature parameter $\tau$ in Eq. (\ref{eq:lossst}) and Eq. (\ref{eq:lossts}) is fixed to $0.05$, and momentum coefficient $\alpha$ in Eq. (\ref{eq:momentumupdate})  is $0.99$.

\begin{table*}[t]
\centering

 \caption{Classification accuracy (mean $\pm$ std $\%$) on \emph{Digits-five} and \emph{DomainNet} datasets.}
 \vspace{-0.1in}
 \resizebox{1.0\textwidth}{!}{
    \begin{tabular}{@{}lcccccc |lccccccc @{}}
    \toprule
    \multicolumn{7}{c|}{\textbf{\emph{Digits-five}}} & \multicolumn{7}{c}{\textbf{\emph{DomainNet}}} \\
    \midrule
    Method & $\rightarrow$ mm& $\rightarrow$ mt
    &$\rightarrow$ up & $\rightarrow$ sv & $\rightarrow$ sy & Avg &  $\rightarrow$ clp& $\rightarrow$ inf
    &$\rightarrow$ pnt & $\rightarrow$ qdr & $\rightarrow$ rel & $\rightarrow$ skt &Avg\\
    \midrule
    Source Only & 63.4$\pm$0.7 & 90.5$\pm$0.8 & 88.7$\pm$0.9 & 63.5$\pm$0.9 & 82.4$\pm$0.6 & 77.7 & 47.6$\pm$0.5  & 13.0$\pm$0.4 & 38.1$\pm$0.5   & 13.3$\pm$0.4 & 51.9$\pm$0.9 & 33.7$\pm$0.5 & 32.9 \\
     MDAN~\cite{zhao2018adversarial} &69.5$\pm$0.3& 98.0$\pm$0.9& 92.4$\pm$0.7& 69.2$\pm$0.6& 87.4$\pm$0.5 & 83.3 &52.4$\pm$0.6& 21.3$\pm$0.8& 46.9$\pm$0.4& 8.6$\pm$0.6& 54.9$\pm$0.6& 46.5$\pm$0.7& 38.4 \\
	 DCTN~\cite{xu2018deep} & 70.5$\pm$1.2 & 96.2$\pm$0.8 & 92.8$\pm$0.3 & 77.6$\pm$0.4 & 86.8$\pm$0.8 & 84.8 &48.6$\pm$0.7 & 23.5$\pm$0.6  &48.8$\pm$0.6  &7.2$\pm$0.5& 53.5$\pm$0.6 & 47.3$\pm$0.5 & 38.2\\
	$\rm M^{3}SDA$~\cite{peng2019moment} & 72.8$\pm$1.1 & 98.4$\pm$0.7 & 96.1$\pm$0.8 & 81.3$\pm$0.9 & 89.6$\pm$0.6 &87.7 &58.6$\pm$0.5& 26.0$\pm$0.9& 52.3$\pm$0.6& 6.3$\pm$0.6& 62.7$\pm$0.5& 49.5$\pm$0.8& 42.6\\	
	 MDDA~\cite{zhao2019multi} &78.6$\pm$0.6& 98.8$\pm$0.4& 93.9$\pm$0.5& 79.3$\pm$0.8& 89.7$\pm$0.7 & 88.1 &59.4$\pm$0.6& 23.8$\pm$0.8& 53.2$\pm$0.6& 12.5$\pm$0.6& 61.8$\pm$0.5& 48.6$\pm$0.8& 43.2\\
    CMSS~\cite{yang2020curriculum} & 75.3$\pm$0.6 & 99.0$\pm$0.1 & 97.7$\pm$0.1 & 88.4$\pm$0.5 & 93.7$\pm$0.2 & 90.8 & 64.2$\pm$0.2 & 28.0$\pm$0.2 & 53.6$\pm$0.4 & 16.0$\pm$0.1 & 63.4$\pm$0.2 & 53.8$\pm$0.4 & 46.5\\	
	 LtC-MSDA~\cite{wang2020learning} & 85.6$\pm$0.8 & 99.0$\pm$0.4 & 98.3$\pm$0.4 & 83.2$\pm$0.6 & 93.0$\pm$0.5 & 91.8 & 63.1$\pm$0.5 & 28.7$\pm$0.7 & 56.1$\pm$0.5 & 16.3$\pm$0.5 & 66.1$\pm$0.6 & 53.8$\pm$0.6 & 47.4\\	
   \midrule
    TCL & \textbf{96.6}$\pm$0.4 & \textbf{99.4}$\pm$0.2 & \textbf{99.3}$\pm$0.2 & \textbf{91.3}$\pm$0.5 & \textbf{97.8}$\pm$0.3 & \textbf{96.9} &\textbf{70.9}$\pm$0.3 & \textbf{30.2}$\pm$0.6 & \textbf{61.8}$\pm$0.4 &\textbf{16.5} $\pm$0.6 & \textbf{72.2}$\pm$0.3 & \textbf{59.6}$\pm$0.4 & \textbf{51.9} \\
    \bottomrule
    \end{tabular}
    }
    \vspace{-0.1in}
    \label{tab:multisource}
\end{table*}

\subsection{Performance Comparison}

\textbf{Single-Source UDA on Office-Home.}
Table \ref{table:results_officehome} reports the classification accuracy comparisons on twelve transfer directions corresponding to \emph{Office-Home} dataset. Generally, our TCL exhibits clear advantages over the existing state-of-the-art methods across all the adaptation directions. Please note that TCL is especially effective and commanding on harder transfer tasks, e.g., Pr $\rightarrow$ Cl and Ar $\rightarrow$ Cl, as the two domains turn out to be substantially different. This might be a good justification that the integration of SSL pretext tasks and domain alignment target in a contrastive genre is a legitimate solution in the context of domain adaptation. Actually, by merely aligning the data distributions between source and target domains, DAN and JAN demonstrate relatively better performance than source only baselines. In comparison, TCL further stresses the impact of cross-domain inner-class invariance while the inter-class discrimination is effectively preserved, and therefore demonstrates stronger resilience against adaptation directions. Please note that RSDA also benefits from pseudo labeling. Nevertheless, RSDA lacks effective mechanism in preserving the discriminativeness across classes and thus performs inferior to TCL. We highlight the method SSDA, which uses instance level rotation prediction as unsupervised learning in addition to the supervised task. SSDA performs drastically worse than our TCL, validating our assumption that instance level self-supervised learning does not necessarily coincide with the optimal parameter region of the supervised task objective.

\textbf{Single-Source UDA on VisDA-2017.}
Table \ref{table:cls-visda} displays the performances of various models on \emph{VisDA-2017}. Similar to the observations on \emph{Office-Home}, TCL again demonstrates strong superiority over its competitors. Particularly, TCL offers significant performance boost on the classes of ``bicycle'', ``person'' and ``truck'' in comparison to the existing state of the art methods. When taking a closer inspection, one might observe that the source only model performs extremely poor on these 3 classes. This well verifies the significant distribution gap between source and target domains on these classes, highlighting the efficiency of our TCL. Moreover, compared to CAN that also exploits pseudo-labeling for domain adaptation, our TCL achieves better performances. The reason might be that CAN updates pseudo labels based on iterative learning during each epoch. Instead, TCL stabilizes pseudo-labeling via a temporally ensembled encoder, and it enforces inner-class invariance among cross-batch instances. Although the source only baseline seemingly offers the best adaptation result on the ``car class'', we conjecture that the model has possibly overfitted to the car class coincidentally in the presence of domain gaps, by sacrificing the bicycle class, person class and etc. From this point, the average score possibly is a more balanced indicator to disambiguate the effectiveness of all algorithms.

\begin{table}[t]
\center
\small
\renewcommand{\arraystretch}{1.1}
\caption{Classification accuracy (mean $\pm$ std $\%$) on \emph{PACS} dataset based on ResNet-18.}
\vspace{-0.1in}
\begin{tabular}{l | c c c c | c}
\hline
Method &$\rightarrow$ Art &$\rightarrow$ Cartoon &$\rightarrow$ Photo &$\rightarrow$ Sketch & Avg \\
\hline
Source Only & 74.9$\pm$0.8 & 72.1$\pm$0.7 & 94.5$\pm$0.5 & 64.7$\pm$1.5 & 76.6 \\
DANN ~\cite{ganin2015unsupervised} & 81.9$\pm$1.1 & 77.5$\pm$1.2 & 91.8$\pm$1.2 & 74.6$\pm$1.0 & 81.5 \\
MDAN ~\cite{zhao2018adversarial} &79.1$\pm$0.3 & 76.0$\pm$0.7 & 91.4$\pm$0.8 & 72.0$\pm$0.8 & 79.6 \\
WBN ~\cite{mancini2018boosting} & 89.9$\pm$0.2 & 89.7$\pm$0.5 & 97.4$\pm$0.8 & 58.0$\pm$1.5 & 83.8 \\
MCD ~\cite{saito2018maximum} &88.7$\pm$1.0 & 88.9$\pm$1.5 & 96.4$\pm$0.4 & 73.9$\pm$3.9 & 87.0 \\
M$^3$SDA ~\cite{peng2019moment} & 89.3$\pm$0.4 & 89.9$\pm$1.0 & 97.3$\pm$0.3 &76.7$\pm$2.8 & 88.3 \\
JiGen ~\cite{carlucci2019domain} & 84.9 & 81.1 & 98.0 &79.1 & 85.7 \\
CMSS ~\cite{yang2020curriculum} & 88.6$\pm$0.3 & 90.4$\pm$0.8 & 96.9$\pm$0.2 & 82.0$\pm$0.5 & 89.5 \\
\hline
TCL & \textbf{93.6}$\pm$0.2 & \textbf{91.7}$\pm$0.3 & \textbf{98.5}$\pm$0.1 & \textbf{87.5}$\pm$0.8 & \textbf{92.8} \\
\hline
\end{tabular}
\vspace{-0.15in}
\label{tab:pacs-da}
\end{table}%

\textbf{Multi-Source UDA on Digits-five, PACS and DomainNet.} Here we compare with several methods on multi-source domain adaptation task: DANN \cite{ganin2015unsupervised}, WBN \cite{mancini2018boosting}, MCD \cite{saito2018maximum},  MDAN \cite{zhao2018adversarial}, DCTN \cite{xu2018deep}, M$^{3}$SDA \cite{peng2019moment}, MDDA \cite{zhao2019multi}, JiGen \cite{carlucci2019domain}, CMSS \cite{yang2020curriculum}, and LtC-MSDA \cite{wang2020learning}. We directly extract the results of the above methods from their publications or from the reports in \cite{wang2020learning, yang2020curriculum}.

Table \ref{tab:multisource} illustrates the performance comparisons on the five multi-source adaption directions of \emph{Digits-five}. Notably, TCL achieves an impressive $\textbf{96.9\%}$ averaged accuracy across 5 directions, significantly surpasses other baselines by a large margin (around $5.1\%$ improvements over the existing best state-of-the-art approach). In addition, TCL also consistently demonstrates distinct advantages in each individual transfer direction, especially on the most challenging direction $\rightarrow$ \textbf{mm} ($96.6\%$), and  $\rightarrow$ \textbf{sv} ($91.3\%$) task. This empirically justifies the advantage of $\mathcal{L}_{tcl-M}$ loss defined in Eq. (\ref{eq:tclM}), where the joint contribution of multiple source domains has been effectively incorporated into the framework of contrastive learning.

As shown in Table \ref{tab:pacs-da}, our TCL achieves state-of-the-art average accuracy of $92.8\%$ on PACS. On the most challenging $\rightarrow$ Sketch task, we obtain $87.5\%$, clearly outperforming other baselines. Note that JiGen is a typical UDA algorithm that utilizes instance level self-supervised learning to improve feature learning. The superiority of TCL over JiGen again justifies advantages of TCL by promoting the instance level assumptions to useful UDA hypothesis.

\begin{table}[t]
\caption{\small The effect of pseudo-labeling loss on target ($\mathcal{L}_{tar}$) and transferrable contrastive loss ($\mathcal{L}_{tcl}$) in our TCL. The mean accuracy over 12 tasks on \emph{Office-Home} and the mean accuracy over $12$ classes on \emph{VisDA-2017} validation set are reported.
}
\vspace{-0.1in}
\begin{center}
\setlength{\tabcolsep}{10pt}
\begin{tabular}{ l  c  c c}
\toprule
Dataset     &  w/o. $\mathcal{L}_{tar}$ &  w/o. $\mathcal{L}_{tcl}$ & TCL\\
\midrule
\emph{Office-Home}  &  72.0 & 71.4 & \textbf{73.4} \\
\emph{VisDA-2017} &  88.7 & 85.5 & \textbf{89.6}  \\
\bottomrule
\end{tabular}
\end{center}
\label{tab:ablation}
\vspace{-0.1in}
\end{table}

Table \ref{tab:multisource} also displays the adaptation performance of various algorithms on \emph{DomainNet}. TCL again outperforms the existing works on most of the adaptation tasks, and TCL achieves the SOTA average accuracy of $51.9\%$. The DomainNet is particularly challenging from two perspectives: Firstly, the domain gap in each adaption direction is significant. Secondly, a relatively large number of categories (i.e., $345$) has made learning discriminative features much more challenging. Nevertheless, TCL successfully relieves these issues, owing to its unique mechanism in the cross-domain class alignment, and TCL correspondingly demonstrates its outstanding ability to find transferable features on this challenging dataset.

\subsection{Experimental Analysis}

\textbf{Ablation Study.} In this section, we investigate the influence of each component in the overall objective $\mathcal{L}_{total-S}$ defined in Eq. (\ref{eq:finallosssingle}). Table \ref{tab:ablation} examines the impact of two key components of TCL: the pseudo-labeling loss on target $\mathcal{L}_{tar}$ and class-specific TCL loss $\mathcal{L}_{tcl}$. By removing $\mathcal{L}_{tcl}$ from $\mathcal{L}_{total-S}$, the averaged accuracy of TCL framework on Office-Home and VisDA-2017 dataset respectively drops by $2.0\%$ and by $4.1\%$. This validates the critical role of $\mathcal{L}_{tcl}$ in the success of TCL, and verifies the importance of cross-domain intra-class alignment. Even if we remove $\mathcal{L}_{tcl}$, TCL still performs better than the comparable RSDA and SE (refer to Table \ref{table:results_officehome} and Table \ref{table:cls-visda}), two algorithms that both capitalize on merely the usage of pseudo label without contrastive loss. This further proves the advantage of our unique strategy of momentum pseudo labeling according to Eq. (\ref{eq:losstarget}).  In addition, if we remove the loss $\mathcal{L}_{tar}$ from Eq. (\ref{eq:finallosssingle}), the performance of TCL also degrades by $1.4\%$ and $0.9\%$. This further supports our assumption that reliable pseudo-label information and class information are critical for the class-level contrastive learning to succeed when embedded in UDA tasks. Without effective learning from the target data, the contrastive loss would be misled by wrong positive pairs, and worsens performance.

\begin{figure}[t]
\center
 \includegraphics[width=0.75\linewidth]{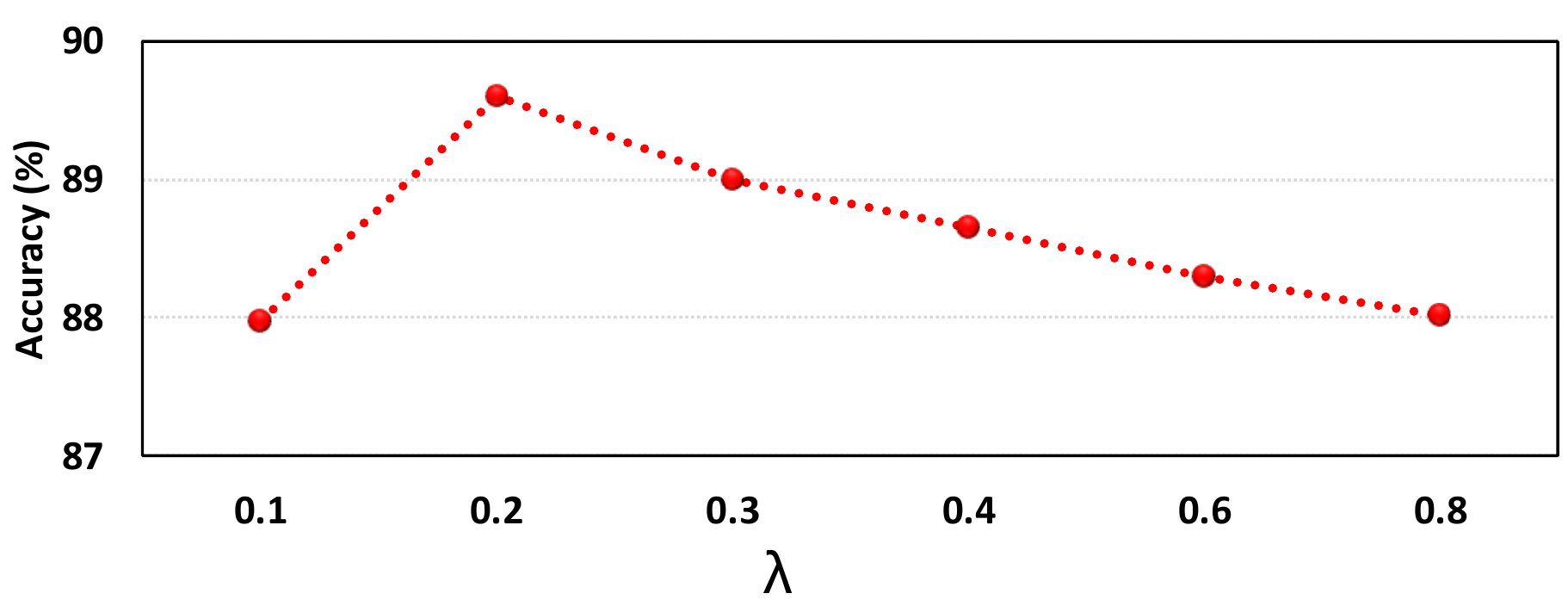}
 \vspace{-0.1in}
 \caption{Effect of hyperparameter $\lambda$ on \emph{VisDA-2017}.}
 \label{fig:sensitivity}
 \vspace{-0.1in}
\end{figure}

\textbf{Sensitivity of Trade-off Parameter $\lambda$.}
We examine the impact of $\lambda (\lambda \in [0,1])$ in Eq. (\ref{eq:finallosssingle}), a hyperparameter that trades off between the classification loss ($\mathcal{L}_{src}$, $\mathcal{L}_{tar}$) and TCL loss $\mathcal{L}_{tcl}$. Figure \ref{fig:sensitivity} shows that TCL model is relatively robust against the change of  $\lambda$, although we observe an evident drop when $\lambda>0.6$. This probably is attributed to the weaker source domain supervision owing to the large $\lambda$, under which the pseudo label becomes less reliable. The pseudo labels therefore becomes more noisy and tend to offer wrong class information to the $\mathcal{L}_{tar}$ and $\mathcal{L}_{tcl}$, which further propagate the error. In contrast, if $\lambda$ is too small (e.g., $\lambda<0.1$), the influence of $\mathcal{L}_{tcl}$ is diluted and TCL loses its unique capability on encouraging cross-domain alignment and inter-class discriminativeness. We observe similar curve pattern on the multi-source adaptation task, and we do not include the result owing to limited space.

\begin{table}[t]
\caption{\small Study on different variants of contrastive learning losses.}
\vspace{-0.0in}
\begin{center}
\small
\setlength{\tabcolsep}{10pt}
\begin{tabular}{ l  c  c c}
\toprule
Dataset     & IDL &  ICDL & TCL\\
\midrule
\emph{Office-Home}  &  70.6 & 72.3 & \textbf{73.4} \\
\emph{VisDA-2017} &  85.1 & 86.7 & \textbf{89.6}  \\
\bottomrule
\end{tabular}
\end{center}
\label{tab:sup}
\vspace{-0.1in}
\end{table}

\textbf{Evaluations of Variants of Contrastive Learning Losses.}
Recall that our proposed TCL loss is a cross-domain class-level contrastive learning objective. To verify the advantage of TCL framework over the conventional instance invariance assumption in the context of UDA, we include two contrastive learning loss variants for comparison.
(1) \textbf{Instance-level Discrimination Loss (IDL):} We replace class-level invariance loss $L_{tcl}$ in Eq. (\ref{eq:tcl}) directly by conventional instance-level contrastive loss on unlabeled target data, i.e., the IDL framework composes of a conventional instance level InfoNCE \cite{oord2018representation} in addition to $L_{src}$ and $L_{tar}$ in Eq. (\ref{eq:finallosssingle}). (2) \textbf{Intra-domain Class-level Discrimination Loss (ICDL):} An alternative to perform class-specific contrastive learning is to replace $L_{tcl}$ by instead penalizing class-level contrastive loss separately for source domain and target domain. We call this framework ICDL. ICDL only considers either intra-source or intra-target class-level discrimination without any cross-domain intra-class constraints. Detailed loss functions of these variants are in supplementary material.

As Table \ref{tab:sup} shows, if we replace $L_{tcl}$ by the IDL loss, the averaged accuracy on Office-Home and VisDA-2017 respectively drops by 2.8\% and by 4.5\% in comparison to TCL framework. The drop well corroborates our hypothesis in the paper that rigid combination of instance level invariance with UDA task is sub-optimal, as the two assumptions does not necessarily coincide. Note IDL loss only emphasizes instance-level discrimination, and thus introduces noise into the memory bank as the negative keys can even be from the same class of each query instance. IDL loss thus inevitably deviates from the motivation of UDA, which aims to erase inter-domain~gap.

Table \ref{tab:sup} also shows the performances of our TCL by replacing ICDL objective for domain adaptation. As expected, none of these modifications has improved over original TCL. The main reason is that ICDL objective does not impose any inter-domain invariance within each class, and therefore cannot effectively erase any domain gap as TCL does.  Please note that our TCL loss goes beyond merely intra-domain class-level instance discrimination as loss ICDL does, and effectively unifies both class-level instance discrimination and inter domain alignment across different domains.

\begin{figure}[t]
\center
 \includegraphics[width=0.78\linewidth]{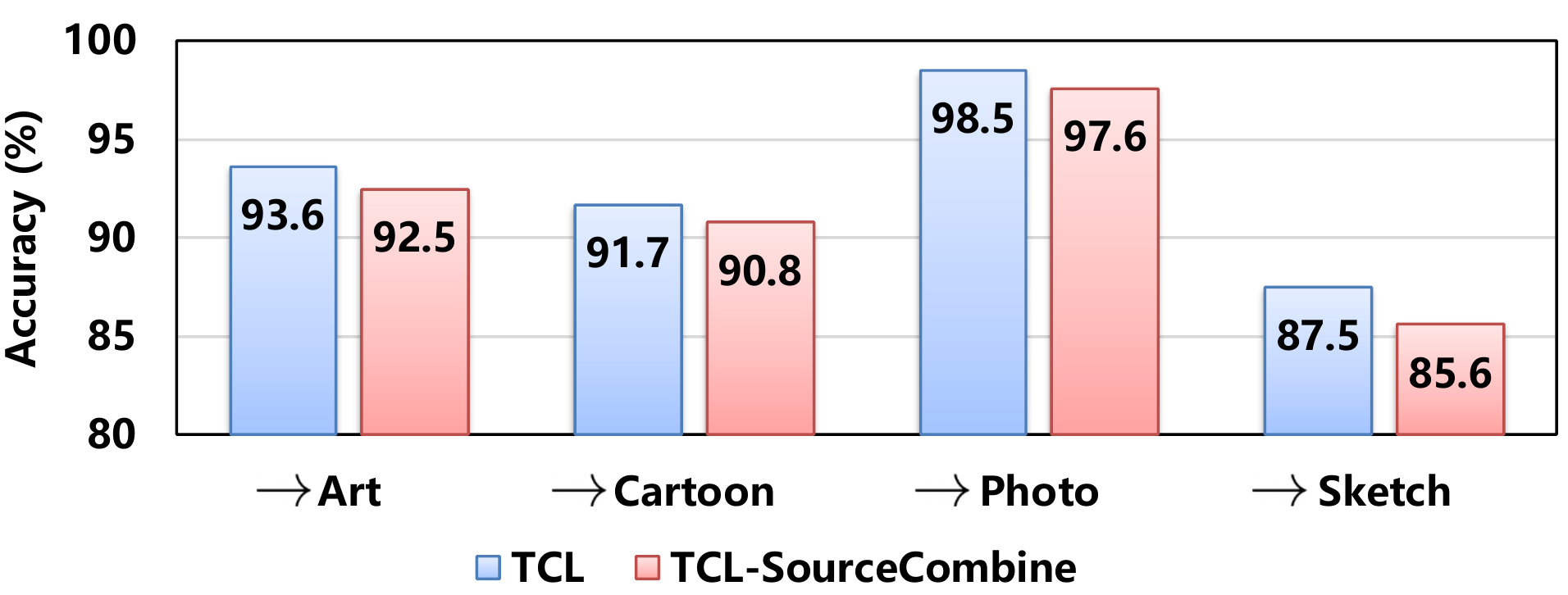}
 \vspace{-0.1in}
 \caption{Comparison with different ways of extending TCL loss to MSDA on PACS.}
 \label{fig:tcl-extending}
 \vspace{-0.2in}
\end{figure}

\textbf{Extending TCL loss to MSDA.} As presented in Eq. (\ref{eq:tclM}), we extend TCL framework to deal with MSDA tasks. Eq. (\ref{eq:tclM}) takes into account the \textbf{\textit{cross source-source domain}} intra-class feature correlations. In comparison, an alternative is to simply combine multiple source domains as a single source domain and directly perform single source TCL loss in Eq. (\ref{eq:tcl}), we call it ``TCL-SourceCombine''.  As shown in Figure (\ref{fig:tcl-extending}), ``TCL-SourceCombine'' is worse than our TCL, justifying that exploring class-level domain alignment across all domains even in the source domain offers further advantage.

\section{Conclusions}

In this work, we present Transferrable Contrastive Learning (TCL), which explores domain adaptation in a self-supervised manner. Particularly, we aim to establish a brand new symbiosis that unifies and fortifies self-supervised pretext tasks (pseudo-labeling and class-level instance discrimination) and domain discrepancy minimization in a single framework. By promoting instance level invariance to inter-domain class-level invariance, TCL elegantly couples the contrastive learning strategy with the goal of UDA. Our proposed TCL loss is devised to integrate both class-level instance discrimination and domain discrepancy minimization via contrastive loss, which best screens the semantic compatibility between query-key pairs across different domains. Our TCL can yield state-of-the-art results on five benchmarks for both single-source and multi-source domain adaptation tasks, especially when compared with the conventional instance level contrastive learning baselines.

\textbf{Acknowledgments.} This work was supported in part by NSFC No. 61872329 and the Fundamental Research Funds for the Central Universities under contract WK3490000005.


\end{document}